\documentclass[11pt]{article}

\usepackage{amsthm, amsmath, amssymb, amsfonts, url, booktabs, tikz, setspace, fancyhdr, amsbsy}
\usepackage[margin = 0.8in]{geometry}
\usepackage{hyperref, enumerate}
\usepackage{esint}
\usepackage{verbatim}
\usepackage{comment}
\usepackage{indentfirst}
\usepackage{graphicx}
\graphicspath{{./images}}
\usepackage{comment}
\usepackage{subcaption}
\usetikzlibrary{math}
\usepackage{placeins}

\newcommand{\xx}{\boldsymbol{x}}

\newcommand{\tth}{\boldsymbol{\theta}}

\begin{document}

\title{Engineering application of physics-informed neural networks for Saint-Venant torsion}

\author{Su Yeong Jo\thanks{Department of Quantum System Engineering, Jeonbuk National University, 567 Baekje-daero, Deokjin-gu, Jeonju 54896, Republic of Korea. \tt{}},\thanks{These authors contributed equally to this work.} 
~Sanghyeon Park\thanks{Department of Mathematics, Inha University, Incheon, Republic of Korea. },
~Seungchan Ko{\footnotemark[3]\tt{}},
~Jongcheon Park\thanks{Electrified Propulsion Structure CAE Team, Hyundai Motor Company, 150 Hyundaiyeonguso-ro, Hwaseong-Si, Gyeonggi-Do 18280 Republic of Korea.},
\\~Hosung Kim{\footnotemark[4]\tt{}},
~Sangseung Lee\thanks{Department of Mechanical Engineering,  Inha University, Incheon, Republic of Korea. \tt{}} \thanks{Corresponding authors.}
~and~Joongoo Jeon\footnotemark[1]\footnotemark[6]\thanks{Graduate School of Integrated Energy-AI, Jeonbuk National University, 567 Baekje-daero, Deokjin-gu, Jeonju 54896, Republic of Korea.}
}
\date{}
\maketitle

\begin{abstract}
The Saint-Venant torsion theory is a classical theory for analyzing the torsional behavior of structural components, and it remains critically important in modern computational design workflows. Conventional numerical methods, including the finite element method (FEM), typically rely on mesh-based approaches to obtain approximate solutions. However, these methods often require complex and computationally intensive techniques to overcome the limitations of approximation, leading to significant increases in computational cost. The objective of this study is to develop a series of novel numerical methods based on physics-informed neural networks (PINN) for solving the Saint-Venant torsion equations. Utilizing the expressive power and the automatic differentiation capability of neural networks, the PINN can solve partial differential equations (PDEs) along with boundary conditions without the need for intricate computational techniques. First, a PINN solver was developed to compute the torsional constant for bars with arbitrary cross-sectional geometries. This was followed by the development of a solver capable of handling cases with sharp geometric transitions; variable-scaling PINN (VS-PINN). Finally, a parametric PINN was constructed to address the limitations of conventional single-instance PINN. The results from all three solvers showed good agreement with reference solutions, demonstrating their accuracy and robustness. Each solver can be selectively utilized depending on the specific requirements of torsional behavior analysis.

\end{abstract}

\section{Introduction}

Torsion problems in mechanical structures are fundamental in both engineering and physics, playing a crucial role in the analysis and design of load-bearing components across numerous industries. Applications range from power transmission systems in automotive and aerospace engineering to mechanical actuators and biomedical devices. A central framework for analyzing such problems is the \textit{Saint-Venant torsion theory}, which provides the basis for computing angular displacement and shear stress distributions in long, prismatic bars subjected to torsional loading. This theory is grounded in elasticity and can be reformulated into a Poisson-type equation using the Prandtl stress function~\cite{ecsedi2010prandtl, ecsedi2009some}, allowing engineers to model torsion under linear elastic assumptions.

For prismatic bars with simple geometries, analytical solutions to the Saint-Venant problem are well established. However, for general cross-sections, especially those that are non-circular or spatially varying, analytical approaches are not feasible. In such cases, numerical methods such as the finite element method (FEM)~\cite{jog2014finite}, finite volume method (FVM)~\cite{chen2019saint}, and spectral techniques like the Galerkin method~\cite{ike2019galerkin} or Fourier-based methods~\cite{ike2023double, ike2024solving} are commonly used. These techniques have been shown to provide accurate solutions even in geometrically complex settings. For instance, Ike and Oguaghamba demonstrated a semi-analytical solution using double finite Fourier sine transforms for rectangular cross-sections~\cite{ike2023double}, while Francu et al. addressed torsion in non-circular bars using a purely analytical formulation~\cite{francu2012torsion}. Nevertheless, the dependency of these approaches on mesh generation and integration grids can lead to significant computational overhead, particularly when dealing with high-resolution 3D domains or multiparametric loading scenarios~\cite{fogang2022cross}.

In recent years, the rise of scientific machine learning has enabled new paradigms for solving partial differential equations (PDEs), notably through the development of \textit{Physics-Informed Neural Networks (PINN)}~\cite{raissi2019physics}. Unlike traditional numerical solvers, the PINN embed the governing physical equations into the loss function of a neural network. This allows for direct approximation of PDE solutions without reliance on mesh generation or discretization, providing a continuous, differentiable representation over the domain of interest. Since their introduction, a number of enhancements to the original PINN formulation have been proposed. An extended PINN (XPINN)~\cite{jagtap2020extended} utilizes domain decomposition strategies to handle large computational domains and improve training convergence. Hierarchical and piecewise variational PINN (hp-VPINN)~\cite{kharazmi2021hp} integrates variational principles and hierarchical mesh refinement strategies to enhance accuracy and efficiency. The auxiliary PINN (A-PINN)~\cite{yuan2022pinn} and a reduced PINN~\cite{nasiri2022reduced} further broaden the range of solvable problems by addressing integro-differential and stiff systems.

While the PINN has demonstrated success in fluid dynamics, heat conduction, and elasticity, its application to torsion problems---especially those governed by the Saint-Venant formulation---remains limited and has not been comprehensively investigated. The inherent complexity of torsion in bars with non-uniform cross-sections, heterogeneous material properties, and non-trivial external torque distributions presents unique challenges. These include sharp gradients in the solution space, high sensitivity to parameter variations, and difficulty achieving stable convergence. Furthermore, most existing PINN models must be retrained for each new configuration of boundary or loading conditions, limiting their utility in design and optimization workflows where rapid evaluation of many cases is required.

In this study, we introduce a tailored PINN framework for solving Saint-Venant torsion problems under a wide range of geometric and loading conditions. Our approach addresses both 1D and 2D formulations, offering a flexible and scalable alternative to mesh-based numerical methods. Initially, we apply the standard PINN to solve the 2D Poisson equation derived from the Prandtl stress function, validating the method across multiple geometries including circular, square, triangular, and irregular domains. We compare the accuracy of the PINN against analytical solutions and FEM benchmarks, demonstrating their ability to capture torsional behavior without requiring mesh generation or numerical integration.

To improve performance in geometrically stiff regions---such as bars with abrupt changes in cross-section or material properties---we adopt the \textit{Variable-Scaling PINN (VS-PINN)} formulation~\cite{ko2025vs}. This method introduces spatial scaling transformations that reduce gradient steepness and improve convergence characteristics. Our results show that the VS-PINN significantly outperform the baseline PINN in both accuracy and training efficiency for challenging 1D torsion cases.

Recognizing the limitations of the standard PINN in parametric studies, we further propose a \textit{Parametric PINN} framework~\cite{cho2024parameterized}. Unlike the conventional PINN, which must be retrained for each new parameter set, the parametric PINN is trained over a range of input parameters---such as torque profile variance or amplitude---allowing for real-time solution inference. This enables fast, mesh-free prediction of torsional response across a wide design space, with applications in design optimization and digital twin systems.

The key contributions of this work are summarized below:
\begin{itemize}
    \item We develop and validate a PINN-based approach for solving the 2D Poisson problem derived from Saint-Venant torsion theory, providing a mesh-free alternative to traditional FEM solvers.
    \item We propose the VS-PINN formulation, which enables efficient and accurate learning in the presence of geometric discontinuities and stiff gradients.
    \item We introduce a parametric PINN framework that generalizes to varying torque conditions, enabling real-time torsional prediction without retraining.
    \item We benchmark all methods against established numerical and analytical solutions, demonstrating competitive accuracy and superior adaptability.
\end{itemize}

Through these contributions, we establish that the PINN---and its advanced variants---can serve as robust, efficient solvers for torsional analysis in engineering structures. The methodology developed here provides a foundation for extending physics-informed learning to broader elasticity problems and reinforces the potential of machine learning in computational mechanics.

\section{Saint-Venant torsion theory}

\subsection{Derivation of Poisson equation}

To solve the Saint-Venant torsion problem using PINNs, it is essential to clearly understand the governing equation. The governing equation must be implemented exactly as the loss function of PINNs. In this section, we derive the Poisson equation governing the Saint-Venant torsion problem. In the Saint-Venant hypothesis, we assumed that the rate of twist is constant and the cross-sections are free to warp in the $z$-direction but the warping is the same for all cross-sections. The displacement components are as follows: 
\begin{equation}\label{displacment component_1}
\begin{aligned}
u &= r\theta(-\sin\beta), &&\\[6pt]
v &= r\theta(\cos\beta), &&\\[6pt]
w &= w(x,y). &&
\end{aligned}
\end{equation}
$u,v,w$ is displacement in the $x,y$ and $z$ directions, respectively. $r$ is radial distance from the torsion axis $r=\sqrt{x^2+y^2}$, $\alpha$ is angle of twist per unit length and $\beta$ is angle defining the position in the cross-section ($\sin\beta=\frac{y}{r}, \cos\beta=\frac{x}{r}$). As the warping is the same for all cross-sections, $w$ is independent in the $z$ direction. Because rate of twist is constant $\frac{{\rm{d}}\theta}{{\rm{d}}z}=\alpha$ $(\theta=\alpha z)$, Eq. \eqref{displacment component_1} can be transformed as follows:
\begin{equation}\label{displacement component_2}
\begin{aligned}
u &= -y\alpha z, &&\\[6pt]
v &= x\alpha z, &&\\[6pt]
w &= w(x,y). &&
\end{aligned}
\end{equation}
With strain-displacement equations, normal strain $\varepsilon$ in the $x,y$ and $z$ can be calculated in Eq. \eqref{abc}.
\begin{equation}\label{abc}
\begin{aligned}
\epsilon_{xx} &= \frac{\partial u}{\partial x} = 0, &&\\[6pt]
\epsilon_{yy} &= \frac{\partial v}{\partial y} = 0, &&\\[6pt]
\epsilon_{zz} &= \frac{\partial w}{\partial z} = 0. &&\\[6pt]
\end{aligned}
\end{equation}
Shear strain $\gamma$ in the $xy$, $yz$ and $xz$ can be calculated in Eq. \eqref{gamma_comp}.
\begin{equation}\label{gamma_comp}
\begin{aligned}
\gamma_{xy}  &= \frac{\partial u}{\partial y} + \frac{\partial v}{\partial x} = 0, &&\\[6pt]
\gamma_{yz}  &= \frac{\partial v}{\partial z} + \frac{\partial w}{\partial y} = \alpha x + \frac{\partial w}{\partial y}, &&\\[6pt]
\gamma_{xz}  &= \frac{\partial u}{\partial z} + \frac{\partial w}{\partial x} = -\alpha x + \frac{\partial w}{\partial x}. &&
\end{aligned}
\end{equation}
In isotropic condition, normal stress $\sigma$ and shear stress $\tau$ are:
\begin{equation*}
\begin{aligned}
\sigma_{xx} &= \sigma_{yy} = \sigma_{zz} = 0, &&\\[6pt]
\tau_{xy}  &= 0, &&\\[6pt]
\tau_{yz}  &= G\gamma_{yz} = G\Bigl(\alpha x + \frac{\partial w}{\partial y}\Bigr), &&\\[6pt]
\tau_{xz}  &= G\gamma_{xz} = G\Bigl(-\alpha y + \frac{\partial w}{\partial x}\Bigr). &&
\end{aligned}
\end{equation*}
$G$ is the shear modulus. Eq. \eqref{equiv} shows the static Equilibrium equation in each direction. Because all shear stresses are independent of the $z$ axis, the equilibrium along the $x$ and $y$ axes is naturally satisfied. 
\begin{equation}\label{equiv}
\begin{aligned}
\frac{\partial \sigma_{xx}}{\partial x} + \frac{\partial \tau_{xy}}{\partial y} + \frac{\partial \tau_{xz}}{\partial z} &= \frac{\partial \tau_{xz}}{\partial z} = 0, &&\\[6pt]
\frac{\partial \tau_{yx}}{\partial x} + \frac{\partial \sigma_{yy}}{\partial y} + \frac{\partial \tau_{yz}}{\partial z} &= \frac{\partial \tau_{yz}}{\partial z} = 0, &&\\[6pt]
\frac{\partial \tau_{zx}}{\partial x} + \frac{\partial \tau_{zy}}{\partial y} + \frac{\partial \sigma_{zz}}{\partial z} &= \frac{\partial \tau_{zx}}{\partial x} + \frac{\partial \tau_{zy}}{\partial } = 0. &&
\end{aligned}
\end{equation}
Prandtl derived a Prandtl (torsion) stress function $\phi(x,y)$ which satisfy the equilibrium equations as follows:
\begin{equation}\label{Prandtl}
\begin{aligned}
\tau_{zx} &= \frac{\partial \phi}{\partial y}, &&\\[6pt]
\tau_{zy} &= -\frac{\partial \phi}{\partial x}. &&
\end{aligned}
\end{equation}
On the other hand, we can derive the stress compatibility equation with Eq. \eqref{equiv} and Eq. \eqref{Prandtl} as follows:
\begin{equation*}
\begin{aligned}
-\frac{\partial \tau_{zy}}{\partial x} + \frac{\partial \tau_{zx}}{\partial y} &= -2G\alpha.
\end{aligned}
\end{equation*}
Using the Prandtl stress function in the stress compatibility equation, the Poisson equation governing the Saint-Venant torsion problem was derived as follows:
\begin{equation*}
\begin{aligned}
-\frac{\partial}{\partial x}\Bigl(-\frac{\partial \phi}{\partial x}\Bigr) + \frac{\partial}{\partial y}\Bigl(\frac{\partial \phi}{\partial y}\Bigr) &= \nabla^2\phi(x,y), &&\\[6pt]
\nabla^2\phi(x,y) &= -2G\alpha. &&
\end{aligned}
\end{equation*}

\subsection{Boundary condition in Saint-Venant torsion theory}

The equilibrium equation in the axial direction is
\begin{equation*}
\begin{aligned}
\tau_{zx}\,{\rm{d}}y\,{\rm{d}}z + \tau_{zy}\,(-{\rm{d}}x\,{\rm{d}}z) &= 0.
\end{aligned}
\end{equation*}
Use Prandtl's stress function(\(\phi_b\)) in boundary :
\begin{equation*}
\begin{aligned}
\frac{\partial \phi_b}{\partial y}\,{\rm{d}}y + \Bigl(-\frac{\partial \phi_b}{\partial x}\Bigr)\,(-{\rm{d}}x) &= 0, &&\\[6pt]
\frac{\partial \phi_b}{\partial y}\,{\rm{d}}y + \frac{\partial \phi_b}{\partial x}\,{\rm{d}}x &= {\rm{d}}\phi_b. &&
\end{aligned}
\end{equation*}
It can be represented in the form of total derivation with respect to \(\phi_b\). Following these two equations, \({\rm{d}}\phi_b=0\), and hence, \(\phi_b\) is constant. Torsional force can be represented in the equation with respect to stress.
\begin{equation*}
\begin{aligned}
T &= \int_A \Bigl( x\,\sigma_{zy} - y\,\sigma_{zx} \Bigr)\,{\rm{d}}A.
\end{aligned}
\end{equation*}
Using Prandtl's stress function, the second boundary condition is: 
\begin{equation*}
\begin{aligned}
T &= 2\int_A \phi(x,y)\, {\rm{d}}A.
\end{aligned}
\end{equation*}
The solution of the derived Poisson equation is essential for calculating the angle of twist based on Saint-Venant torsion theory. However, in conventional numerical methods, as the complexity of the geometry increases, a finer mesh is required, leading to a significant increase in computational cost due to both the expense of mesh generation and the cost of matrix operations (the curse of dimensionality). Compared to traditional numerical methods, the PINN is advantageous in handling high-dimensional computations. Therefore, this study compares the performance of PINN and conventional numerical methods in solving the Poisson equation.

\section{Physics-informed neural networks
}

\begin{figure}
    \centering
    \begin{subfigure}{0.6\textwidth}
        \centering
        \includegraphics[width=\linewidth]{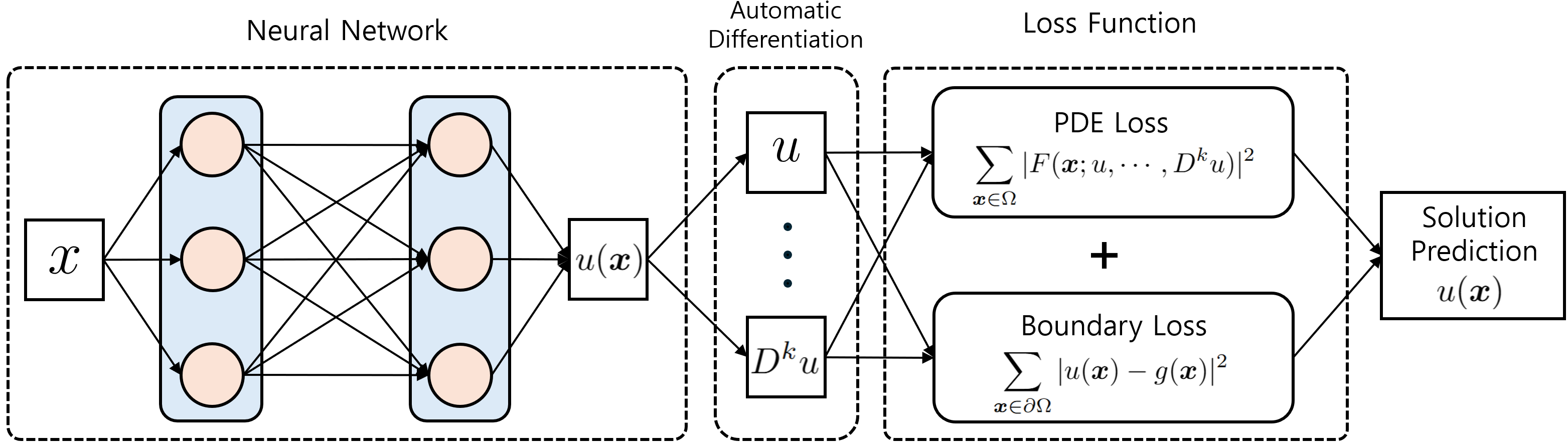}
        \caption{Standard PINN}
        \label{spinn}
    \end{subfigure}    
    \begin{subfigure}{0.6\textwidth}
        \centering
        \includegraphics[width=\linewidth]{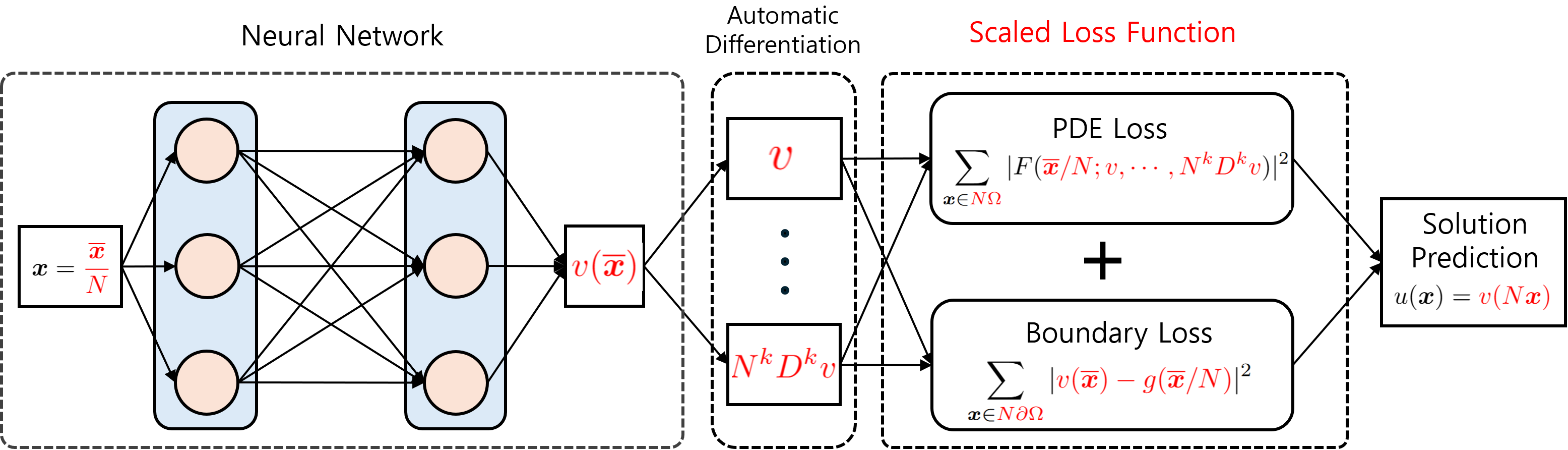} 
        \caption{VS-PINN}
        \label{vspinn}
    \end{subfigure}

    \begin{subfigure}{0.6\textwidth}
        \centering
        \includegraphics[width=\linewidth]{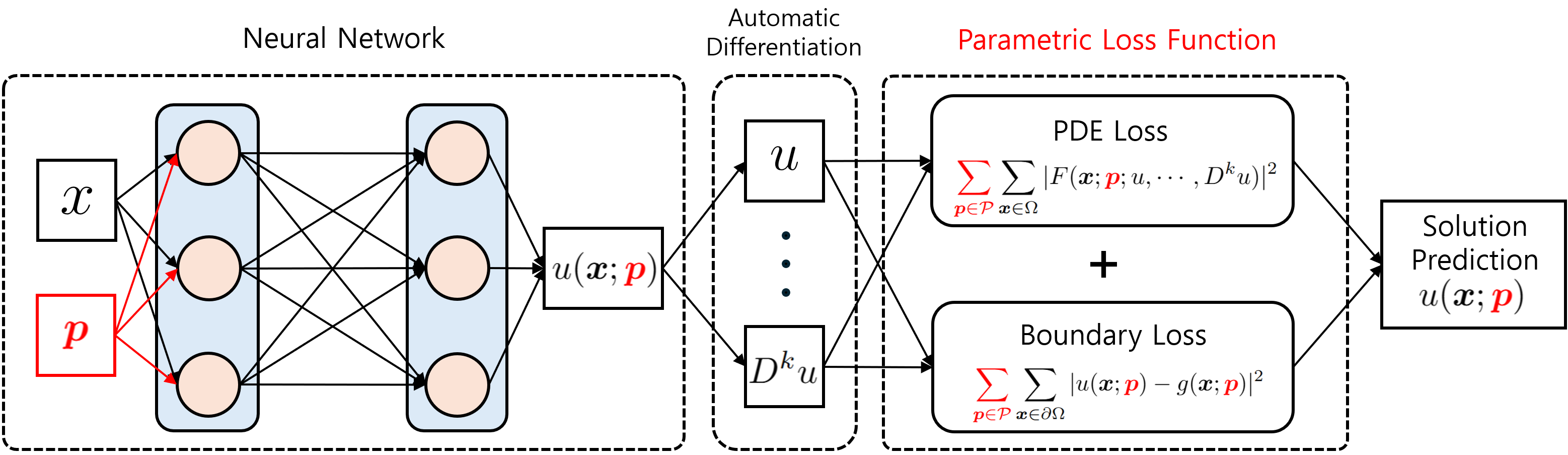} 
        \caption{Parametric PINN}
        \label{ppinn}
    \end{subfigure}
\caption{Comparison of schematic diagrams of PINN, VS-PINN and Parametric PINN.}
\label{comp_pinn}
\end{figure}

In this section, we shall provide a brief introduction to the PINN framework along with several more advanced variants, which will be utilized in the later sections. We begin by presenting the standard PINN, which serves as the baseline model \cite{pinn01}. Next, we describe a novel approach called VS-PINN, which significantly enhances the training efficiency of the PINN \cite{vspinn}. As we will demonstrate later, this method leads to substantial improvements in both accuracy and training efficiency when applied to our problem. Finally, we will propose the parametric PINN, an advanced method that enables the rapid prediction of solutions without retraining when parameters determining the PDE problem, such as boundary conditions, initial conditions, or external forces, are varied. A comparative summary of these three approaches is provided in Figure \ref{comp_pinn}.

\subsection{Vanilla PINN}

In this section, we provide a brief summary of a PINN \cite{pinn01, pinn02}. For the description of this method, let us consider the following model problem defined on a bounded domain $\Omega\subset\mathbb{R}^d$: \begin{equation}\label{main_eq} 
\begin{aligned} \mathcal{D}[u] &= f(\xx), && \boldsymbol{x} \in \Omega,\\ u(\boldsymbol{x}) &= g(\boldsymbol{x}), && \boldsymbol{x} \in \partial\Omega, \end{aligned} \end{equation} where $\mathcal{D}$ is a differential operator and $u:{\Omega} \to \mathbb{R}$ is the unknown, with the (possibly multi-dimensional) variable $\xx = (x_1, x_2, \dots, x_d) \in \Omega$. For simplicity, we treat the time variable $t$ as an additional coordinate of $\xx$, which allows $\Omega$ to represent a spatio-temporal domain. Consequently, an initial condition can be considered as a particular type of boundary condition on $\partial\Omega$.

The main concept of the PINN is to approximate the solution $u(\xx)$ with a neural network $u(\xx; \tth)$, where $\tth$ represents the parameters of the neural network. The network is trained by minimizing the so-called {\textit{physics-informed loss function}}, which combines the residual of the PDE over the domain with the discrepancies between the neural network and the boundary conditions. To be more specific, the network is trained to minimize the total loss function, defined as \begin{equation}\label{total_loss} 
\mathcal{L}_{\rm total}(\tth) = \lambda_r \mathcal{L}_{ r}(\tth) + \lambda_b \mathcal{L}_{ b}(\tth), 
\end{equation} 
where the residual loss and the boundary loss functions are given by: \begin{equation}\label{res_loss} \mathcal{L}_{ r}(\tth) = \frac{1}{N_r} \sum_{i=1}^{N_r} \left| \mathcal{D}[u](\boldsymbol{x}_r^i; \tth) - f(\boldsymbol{x}_r^i) \right|^2\quad{\rm{and}} \quad \mathcal{L}_{b}(\tth) = \frac{1}{N_b} \sum_{j=1}^{N_b} \left| u(\boldsymbol{x}_b^j; \tth) - g(\boldsymbol{x}_b^j) \right|^2. \end{equation} Here, $N_r$ and $N_b$ are the batch sizes, and $\lambda_r$ and $\lambda_b$ are the weights associated with the residual and boundary terms respectively. A typical strategy is to randomly sample $\boldsymbol{x}_r^i$ and $\boldsymbol{x}_b^j$ from uniform distributions $\mathcal{U}(\Omega)$ and $\mathcal{U}(\partial\Omega)$ respectively. All the derivatives involved for the loss functions can be computed effectively using {\textit{reverse-mode automatic differentiation}} \cite{auto_diff_1, auto_diff_2}. This ensures that the boundary condition is encoded, while the physical laws expressed as PDEs are incorporated over the computational domain. This is the main reason the method is called a `physics-informed' neural network. Moreover, if sensor data $(\xx_s^i, y_s^i)$, where $y_s^i = u(\xx_s^i)$ for $i = 1, 2, \dots, N_s$, is available for training (i.e. if the ground-truth values on $\{\xx^i_s\}_{i=1}^N$ are known), an extra supervised loss term $\mathcal{L}_{s}(\tth) = \sum_{i=1}^{N_s} \left| u(\xx_s^i; \tth) - y_s^i \right|^2$ can be added to the total loss function. This additional term typically enhances the model's performance. However, because our objective is to develop PINN as a solver for the Saint-Venant torsion problem, the ground truth was not used. The schematic diagram of PINNs can be found in Figure \ref{spinn}.

\subsection{Variable-scaling PINN}\label{sec:VS-PINN}
In this section, we introduce the recently developed method for training PINN \cite{vspinn}, which can improve the training process. This new method exploits a variable-scaling technique during the training process. To be more specific, we utilize the variable transformation $\xx \mapsto \overline{\xx} / N$ to define the loss function, aiming to mitigate the potential steep behavior of solutions. To further elaborate on the methodology, let us write the given PDE problem in an implicit manner: 
\begin{equation}\label{given_PDE}
\begin{aligned}
    F(\xx; u, Du, D^2u, \dots, D^k u) &= 0, \quad && \xx \in \Omega, \\
    u(\xx) &= g(\xx), \quad && \xx \in \partial\Omega,
\end{aligned}
\end{equation}
where the function $F(\xx)$ is defined as $F(\xx) = \mathcal{D}[u](\xx) - f(\xx)$ in accordance with the notation Eq. \eqref{main_eq}, and $D^j$ represents the mixed partial derivatives of order $j \in \mathbb{N}$. The corresponding loss function for the PINN is then expressed as:
\begin{equation}\label{PINN_loss}
    \mathcal{L}_{\rm original}
    = \frac{\lambda_r}{N_r} \sum_{i=1}^{N_r} \bigg|F(\xx^i_r; u(\xx^i_r), Du(\xx^i_r), D^2u(\xx^i_r), \dots, D^k u(\xx^i_r))\bigg|^2+ \frac{\lambda_b}{N_b} \sum_{j=1}^{N_b} \left|u(\xx^j_b) - g(\xx^j_b)\right|^2.
\end{equation}
Here, the points $\xx^i_r$ and $\xx^j_b$ are independently and identically distributed (i.i.d.) random samples drawn from uniform distributions, specifically $\xx^i_r \sim \mathcal{U}(\Omega)$ and $\xx^j_b \sim \mathcal{U}(\partial\Omega)$.  

Next, we introduce a new variable transformation $\overline{\xx} = N\xx$, which maps the original domain $\Omega$ to the rescaled domain $N\Omega := \{N\xx : \xx \in \Omega\}$. Using this notation, we define $u(\overline{\xx}/N) = v(\overline{\xx})$. Intuitively, the transformation $\xx = \overline{\xx}/N$ can be interpreted as a `zoom-in' on the solution profile. This scaling effect helps to smooth out sharp variations in the solution dynamics over the domain. The computation of the twist angle in the shaft geometry featuring a diameter transition serves as a prototypical example of a sharp variation. By applying the chain rule, we obtain
\[
D_{\xx} u(\xx) = D_{\overline{\xx}} u\left(\frac{\overline{\xx}}{N}\right) \frac{{\rm d}\overline{\xx}}{{\rm d}\xx} = N D_{\overline{\xx}} v(\overline{\xx}),
\]
and this result extends naturally to higher-order derivatives:
\[
D^k_{\xx} u(\xx) = N^k D^k_{\overline{\xx}} v(\overline{\xx}).
\]
Thus, the original PDE in Eq. \eqref{given_PDE} can be reformulated as follows:
\begin{equation}\label{rewritten_PDE}
\begin{aligned}
    F(\overline{\xx}/N;v,NDv,N^2D^2v,\cdots,N^kD^kv)&=0 &&\overline{\xx}\in N\Omega,\\
    v(\overline{\xx})&=g(\overline{\xx}/N) && \xx\in N\partial\Omega.
\end{aligned}
\end{equation}
The corresponding loss function for the reformulated equation Eq. \eqref{rewritten_PDE} is then defined as
\begin{equation}\label{scaled_loss}
    \mathcal{L}_{\rm scaled}
    =\frac{\lambda_r}{N_r}\sum^{N_r}_{i=1}\bigg|F\bigg(\frac{\overline{\xx}^i_r}{N};v(\overline{\xx}^i_r),NDv(\overline{\xx}^i_r),N^2D^2v(\overline{\xx}^i_r),\cdots,N^kD^kv(\overline{\xx}^i_r)\bigg)\bigg|^2+\frac{\lambda_b}{N_b}\sum^{N_b}_{j=1}\bigg|v(\overline{\xx}^j_b)-g\bigg(\frac{\overline{\xx}^j_b}{N}\bigg)\bigg|^2. 
\end{equation}
Here, the points $\overline{\xx}^i_r$ and $\overline{\xx}^j_b$ are identically and independently drawn from uniform distributions $\overline{\xx}^i_r \sim \mathcal{U}(N\Omega)$ and $\overline{\xx}^j_b \sim \mathcal{U}(N\partial\Omega)$ respectively. In this rescaled loss function, the weighting parameters $\lambda_r$ and $\lambda_b$ must be carefully selected based on the effects of scaling. For instance, they can be adjusted to eliminate common factors or mitigate extreme coefficients. The key idea of this method is that we minimize the scaled loss function Eq. \eqref{scaled_loss} instead of the original formulation Eq. \eqref{PINN_loss}. Once training is completed and we obtain the predicted solution $v(\overline{\xx})$, we revert the scaled variable to the original variable using the transformation $u(\xx) = v(N\xx)$. This yields the desired solution in the original problem setting. Since the proposed method expands the computational domain from $\Omega$ to $N\Omega$, one might wonder whether additional random samples are needed for training. However, as shown in the numerical experiments in \cite{vspinn}, this is not necessary. In fact, we can achieve a significant improvement in training efficiency while using the same number of random samples, even from the scaled domain. The graphical summary of the VS-PINN can be found in Figure \ref{vspinn}.

\subsection{Parametric PINN}\label{sec:Pa_PINN}
Although the PINN have successfully applied to a wide range of engineering problems, the standard PINN can handle only a single instance at a time; therefore, any change in the PDE data such as boundary conditions, initial conditions, or external forces, necessitates retraining the model from scratch. The resolution of this single‑instance challenge is essential for the development of a solver for the Saint–Venant torsion problem. To overcome this limitation, we shall propose a parametric PINN approach \cite{kaltenbach2022deep, para_pde}, and we will explore a methodology that can make real-time solution predictions for varying PDE data by incorporating a new technique into the standard PINN framework. This approach is conceptually similar to recent operator learning-based methods, such as DeepONet \cite{DeepOnet} and the Fourier Neural Operator \cite{fno}. These methods are known to perform well on parametric problems when a sufficiently large amount of training data is available. However, they require extensive training datasets composed of input-output pairs, which are often costly to generate, and their large and complex architectures result in a substantial computational burden during training. In contrast, our method—while potentially achieving slightly lower final accuracy—requires no training data and employs a much simpler architecture, offering significant advantages in terms of computational efficiency.

For a parameter space $\mathcal{P}\subset\mathbb{R}^m$ for some $m\in\mathbb{N}$, let $\boldsymbol{p}\in{\mathcal{P}}$ be the parameter that determines the PDE data. As an example, as will be utilized in the later section, the parameter $\boldsymbol{p}=(p_1,p_2,p_3)$
can characterize the Gaussian distribution function
\[
    f(x;\boldsymbol{p})= p_1\exp\bigg(-\frac{(x-p_2)^2}{2|p_3|^2}\bigg).
\]
The core idea of the parametric PINN lies in treating the parameters that determine the data as additional inputs to the neural network. By doing so, training is performed not only on the physical domain $\Omega$ but also on the parametric domain $\mathcal{P}$. This enables the parametric PINN to rapidly predict solutions corresponding to both a specific spatial location and a particular set of parameters. Let us further elaborate on the idea. For each parameter $\boldsymbol{p}\in{\mathcal{P}}$, the original PINN aims to solve the minimization problem
\[
\min_{\tth} \mathcal{L}_{\rm{total}}(\tth; \boldsymbol{p}),
\]
where $\mathcal{L}_{\rm{total}}(\tth;\boldsymbol{p})$ is the physics-informed loss function defined in Eq. \eqref{total_loss} corresponding to the fixed parameter $\boldsymbol{p}\in{\mathcal{P}}$. The approach for solving parametric PDE problems with PINNs is to use deep neural networks which take both $\boldsymbol{p}\in\mathcal{P}$ and $\xx\in\Omega$ as inputs. Consequently, we need to consider the following form of population loss function for a parametric PINN:
\[
\mathcal{L}_{\rm{para}}(\tth):=\int_{\mathcal{P}}\mathcal{L}_{\rm{total}}(\tth;\boldsymbol{p})\,{\rm{d}}\mu(\boldsymbol{p})
\]
for some suitable measure $\mu$ on $\mathcal{P}$. In practice, for a given parameter space $\mathcal{P}\subset \mathbb{R}^m$, we consider the following parametric empirical loss functions
\begin{equation}\label{parametric_loss}
\begin{aligned}
 \mathcal{L}_{{\rm para},\,r}(\tth) &= \frac{1}{N_{pr}}\frac{1}{N_r} \sum_{k=1}^{N_{pr}}\sum_{i=1}^{N_r} \left| \mathcal{D}_{\boldsymbol{p}^k}[u](\boldsymbol{x}_r^i;\boldsymbol{p}_r^k; \tth) - f(\boldsymbol{x}_r^i;\boldsymbol{p}_r^k) \right|^2, \\ 
 \quad \mathcal{L}_{{\rm para},\,b}(\tth) &= \frac{1}{N_{pb}}\frac{1}{N_b} \sum_{\ell=1}^{N_{pb}}\sum_{j=1}^{N_b} \left| u(\boldsymbol{x}_b^j;\boldsymbol{p}_b^\ell; \tth) - g(\boldsymbol{x}_b^j;\boldsymbol{p}_b^\ell) \right|^2. 
\end{aligned}
\end{equation}
Here, the points $\xx^i_r$, $\xx^j_b$, $\boldsymbol{p}^k_r$ and $\boldsymbol{p}^\ell_b$ are i.i.d. random samples drawn from uniform distributions $\xx^i_r \sim \mathcal{U}(\Omega)$, $\xx^j_b \sim \mathcal{U}(\partial\Omega)$ and $\boldsymbol{p}^k_r,\boldsymbol{p}^\ell_b \sim \mathcal{U}(\mathcal{P})$. To be more specific, $\mathcal{D}_{\boldsymbol{p}^k}$ is a differential operator, $f(\cdot;\boldsymbol{p}^k)$ and $g(\cdot;\boldsymbol{p}^\ell)$ are external forces and boundary conditions respectively, all determined by the parameter $\boldsymbol{p}\in\mathcal{P}$. Finally, we obtain the solution prediction $u(\cdot,\cdot;\tth):\mathbb{R}^{\dim(\Omega)+\dim(\mathcal{P})}\rightarrow \mathbb{R}$ by minimizing the parametric total loss defined as
\[\mathcal{L}_{\rm para}(\tth) = \lambda_r \mathcal{L}_{{\rm para},\,r}(\tth) + \lambda_b \mathcal{L}_{{\rm para},\,b}(\tth) \]
for appropriately chosen weights $\lambda_r$, $\lambda_d>0$.
After training is completed, we can obtain the solution whose input is both the spatial variable $\xx\in\Omega$ and the parametric variable $\boldsymbol{p}\in\mathcal{P}$, so that we can make a real-time solution prediction for varying parameter $\boldsymbol{p}\in\mathcal{P}$ within a physically meaningful range for engineering applications. The schematic diagram of the parametric PINN is depicted in Figure \ref{ppinn}.

\section{Applications}
\subsection{2D Poisson equation with scaled Prandtl stress function}
\begin{figure}[h]
    \centering
    \includegraphics[width=\linewidth]{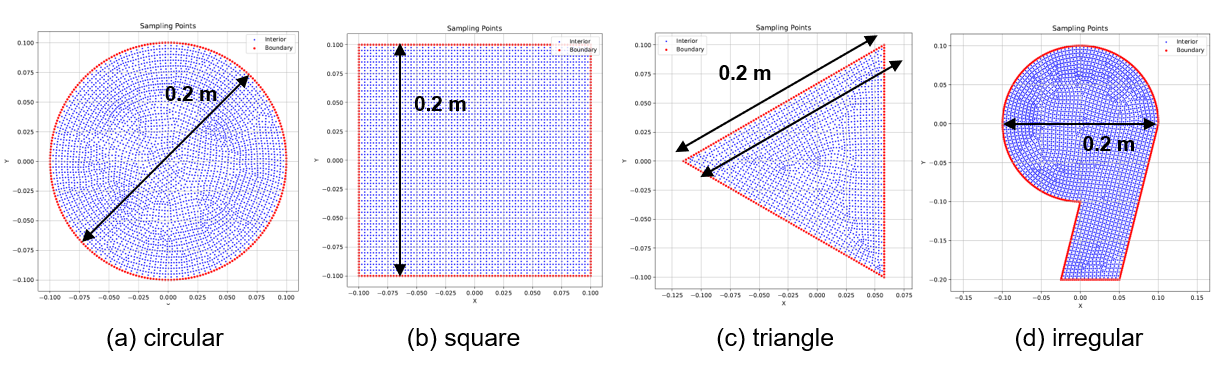}
    \caption{Case study of four shapes. The Poisson equation with the Dirichlet boundary condition is solved.}
    \label{var_shapes}
\end{figure}

\;Torsion constant is a kind of polar moment of inertia and is required to calculate the torsional stiffness of a cross-section. The torsional constant is a challenging constant to compute. Its calculation depends on the distribution of shear stress, which in turn is highly dependent on the cross-sectional shape. Therefore, if the shear stress distribution function is provided, the torsional constant for a unit rotation can be obtained by integrating over the entire cross-section. However, as mentioned above, since the shear stress distribution greatly depends on the cross-sectional shape and varies significantly from one section to another, there is no single equation or formula that covers all cases. Consequently, various numerical methods are used to determine the torsional constant. Calculating the torsional constant for irregular polygonal shapes is an extremely challenging task.
\begin{equation*}
\begin{aligned}
J &= \frac{2}{G\alpha}\int_A \phi(x,y)\, {\rm{d}}A.
\end{aligned}
\end{equation*}
Since $G$ and $\alpha$ are constants, we can obtain the torsional constant by finding the stress function field by computing the Poisson equation. In this study, for more straightforward calculations, we define a scaled Prandtl stress function $\phi'=\phi/\alpha$, and find $\phi'$ from the Poisson equation as shown in Eq. \eqref{poi_G}.
\begin{equation}\label{poi_G}
\begin{aligned}
\nabla^2\phi'(x,y) &= -2G.
\end{aligned}
\end{equation}
Integrating $\phi'$ over the cross-section and multiplying by $2/G$ gives the torsional constant as follows:
\begin{equation*}
\begin{aligned}
J &= \frac{2}{G}\int_A \phi'(x,y)\, {\rm{d}}A.
\end{aligned}
\end{equation*}
In this paper, we solve $\phi'$ using PINN for the cross sections of a circle with a diameter of 0.2m, an equilateral triangle with 0.2 m, a square with 0.2m, and an irregular shape of similar scale (Figure \ref{var_shapes}). 
In this study, we used the PINN to solve $\phi'$, as its output. The loss function for the PINN was designed to minimize the residual of the Poisson equation as shown in Eq. \eqref{poi_loss}. For the boundary condition, the loss function is based on the fact that, according to Saint-Venant torsion, $\phi'$ must always satisfy a constant value at the boundary. For computational convenience, $\phi'$ is assumed to be 0. This assumption is used to define the boundary condition loss function. The final loss function is the sum of the loss function for the PDE and the boundary condition, multiplied by weighting factors $\lambda_r=1$ and $\lambda_b=100,000$. In this study, the performance of PINN was compared with ANSYS software, and the sampling points of PINN and the ANSYS grid were set to be the same.
\begin{equation}\label{poi_loss}
\begin{aligned}
\mathcal{L}_{r} &= \nabla^2\phi'(x,y) + 2G, &&\\[6pt]
\mathcal{L}_{b}  &= (\phi'_{bc})^2, &&\\[6pt]
\mathcal{L}_{\rm{total}} &= \lambda_r\mathcal{L}_{r} + \lambda_b\mathcal{L}_{b}. &&
\end{aligned}
\end{equation}

\subsection{1D Poisson equation with angle of twist}

If we can calculate the torsional constant as above, it becomes possible to calculate the angle of twist in the shaft. With the torsional constant, the boundary condition on the edge can be expressed as follows:
\begin{equation}\label{1D_torsion_equ_derv_1}
\begin{aligned}
    T=GJ\alpha=GJ\frac{{\rm{d}}\theta}{{\rm{d}}x},\\
    \frac{{\rm{d}}}{{\rm{d}}x}\left(GJ\frac{{\rm{d}}\theta}{{\rm{d}}x}\right)=0.
\end{aligned}
\end{equation}
If the torsional constant of the shaft changes along the x-axis, the equation can be expressed as follows:
\begin{equation}\label{1D_torsion_equ_derv}
\begin{aligned}
    \frac{{\rm{d}}}{{\rm{d}}x}\left(GJ(x)\frac{{\rm{d}}\theta}{{\rm{d}}x}\right)=0.
\end{aligned}
\end{equation}
In this study, we evaluated and improved the PINN performance under shaft conditions with a stiff changing diameter.

\section{Results and Conclusions
}

\subsection{ANSYS software for torsional constant
}

\begin{figure}[h]
    \centering
    \includegraphics[width=\linewidth]{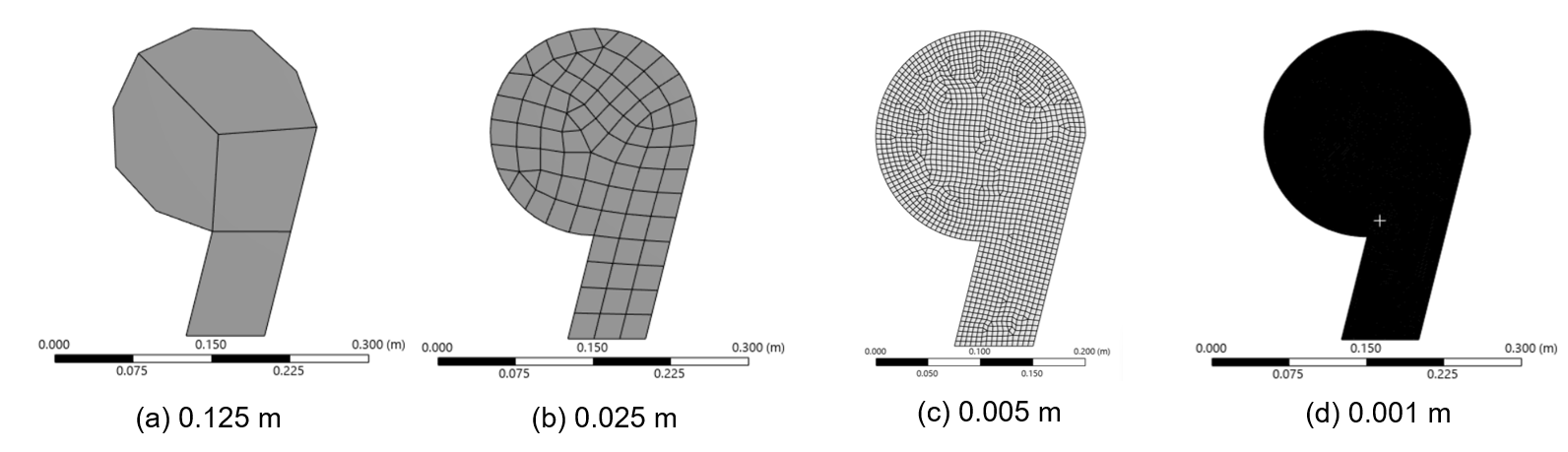}
    \caption{Sensitivity study for grid size determination. The unified grid size was determined to be 0.005 m.}
    \label{sensi_st}
\end{figure}

In order to determine a unified grid size for all geometries, the required minimum grid size was investigated based on the most complex geometry. As shown in Figure \ref{sensi_st}, a grid sensitivity study was performed. Based on an average $\phi'$ value of 0.001 m, a relative error of 61.6\%, 5.65\% and 2.29\% was observed at 0.125, 0.025 and 0.005 m grid size. The error of 2.29\% was considered acceptable, and hence grid size was determined to be 0.005 m.

\begin{figure}[h]
    \centering
    \includegraphics[width=\linewidth]{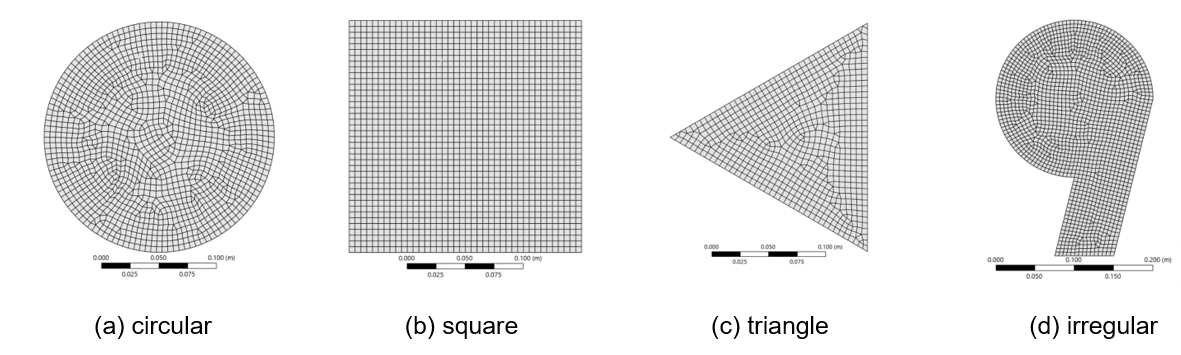}
    \caption{0.005 m grid modeling for each shape. The spatial coordinates of the grid generated in ANSYS were used identically as the sampling points of PINN.}
    \label{small_grid}
\end{figure}

As shown in Figure \ref{small_grid}, 0.005 m grids were constructed using ANSYS software for all geometries. The square domain is constructed with a uniform structured mesh, while the circular, triangle and irregular geometries employ an unstructured mesh that offers high adaptability to complex shapes. The Poisson equation with the Dirichlet boundary condition was solved. The ANSYS results were compared with the PINN results in Section \ref{sec:PINN_compara}.


\subsection{PINN for torsional constant
}\label{sec:PINN_compara}

\begin{figure}[h]
    \centering
    \includegraphics[width=\linewidth]{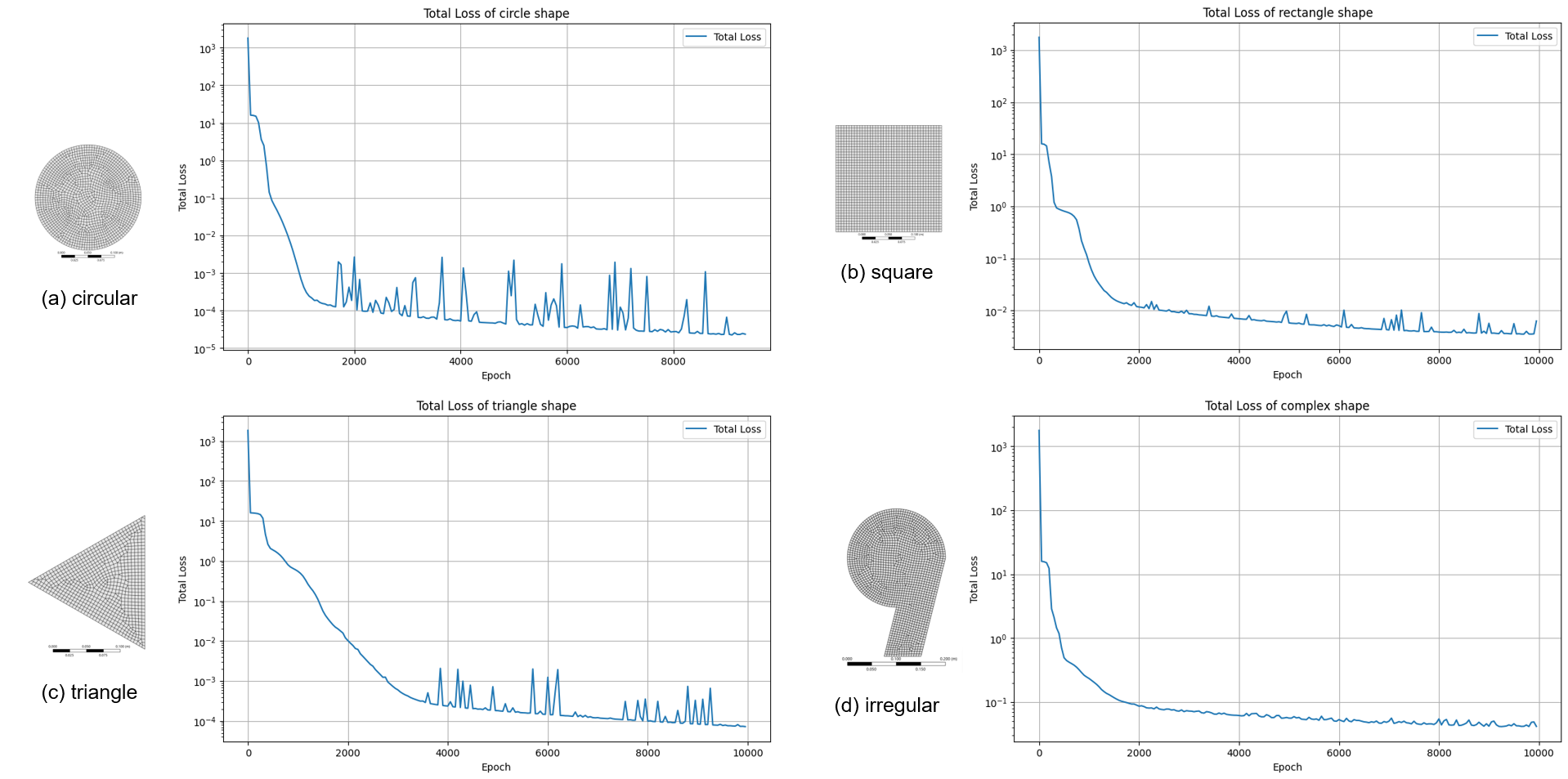}
    \caption{Training loss by epoch. Training was terminated when the loss converged in all shapes.}
    \label{shape_loss}
\end{figure}

For each shape, PINN calculated the scaled Prandtl stress function under the sampling point and training loss conditions described above. Figure \ref{shape_loss} shows the training loss curves of PINN for different geometries: circle, rectangle, triangle, and an irregular shape. Across all shapes, the total loss decreases rapidly in the early epochs, indicating effective learning of the governing physics. While the rectangle and irregular shapes appear to converge more smoothly, the circle and triangle shapes achieve significantly lower loss values—reaching scales as low as $10^{-5}$ to $10^{-4}$. Training was terminated when the loss converged (about 10,000 epochs). 

\begin{figure}[h]
    \centering
    \includegraphics[width=\linewidth]{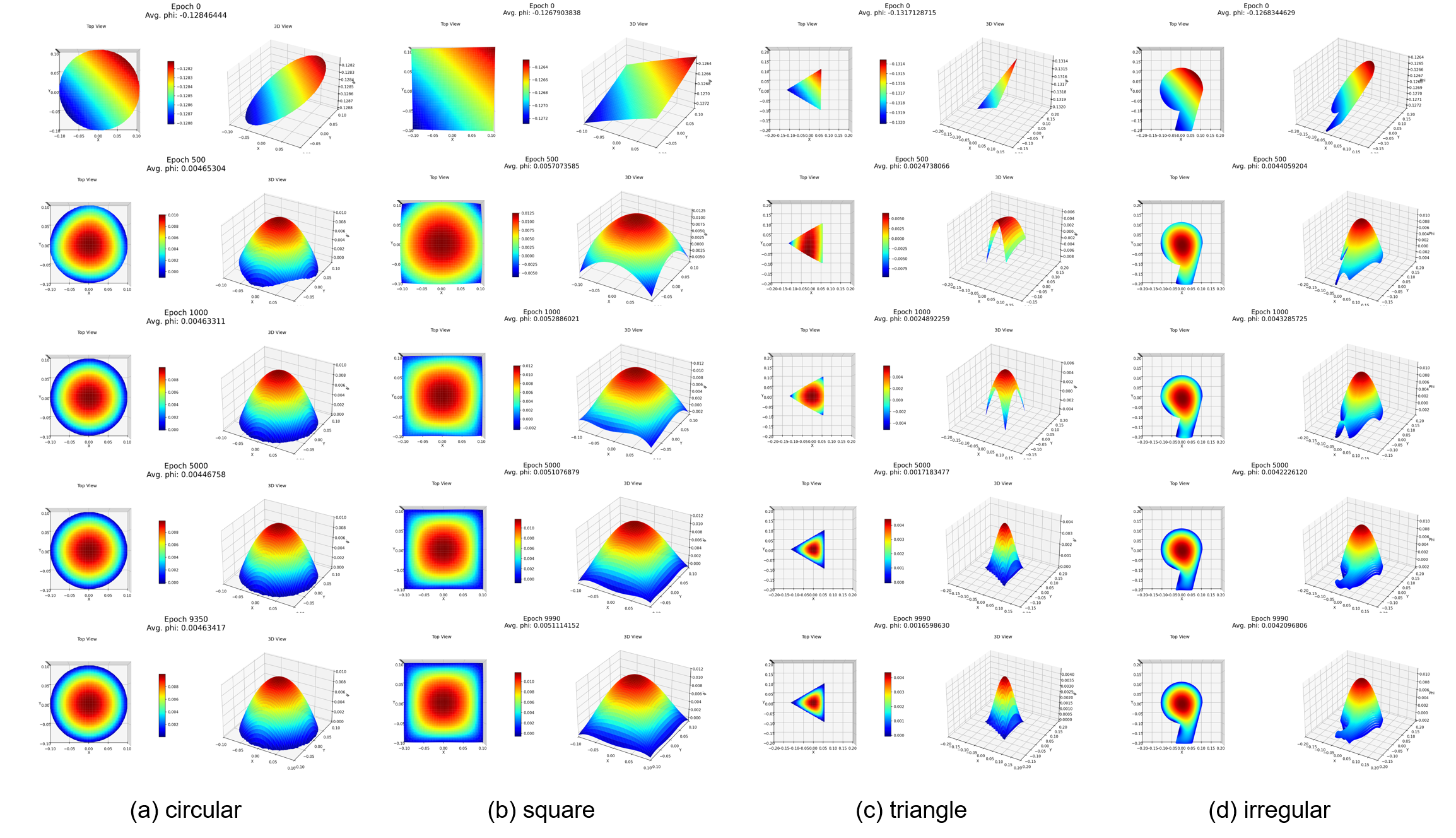}
    \caption{Prandtl stress function field over epochs. It was observed that the field converged as learning progressed.}
    \label{learning_3D}
\end{figure}

Figure \ref{learning_3D} visualizes the learning progression of the PINN by showing the predicted contour and 3D surface plots of the solution $\phi'$ across different epochs for each geometry. At epoch 0, the predictions are far from the true solution, with large deviations. As training progresses (epochs 500 to 9990), all geometries gradually converge to a smooth and symmetric solution, consistent with the expected parabolic profile governed by the PDE. The circle and square shapes exhibit faster and more stable convergence to the true solution, maintaining symmetry early on. In contrast, the triangle and complex shapes take longer to stabilize, but eventually achieve accurate contours.

Table \ref{tab:summary} compares the results from ANSYS, PINN, and the analytical solution with relative errors (compared to the analytical solution) shown in parentheses. For the circular, square, and triangle shapes, analytical solutions are available, enabling error comparison. The results show that the PINN predictions are remarkably accurate, with relative errors of only 0.1\% for both the circular and square shapes, and 3.0\% for the triangle. In contrast, ANSYS exhibits noticeably higher errors—5.5\%, 6.2\%, and 10.4\%, respectively. It is important to note that the PINN demonstrates superior performance under the same mesh or sampling point resolution. For the irregular shape, where no analytical solution exists, only the absolute values are reported, with PINN again showing slightly lower values than ANSYS. Overall, this table highlights the potential of PINN for accurate solution prediction across various geometries.

\begin{table}[h]
\centering
\resizebox{\textwidth}{!}{%
\begin{tabular}{lccc}
\textbf{Shape} & \textbf{ANSYS} & \textbf{PINN} & \textbf{Analytical Solution} \\
\hline
circular  & 4.88690E-03 (5.5\%)  & 4.63417E-03 (0.1\%) & 4.63202E-03 \\
square    & 5.43650E-03 (6.2\%)  & 5.11415E-03 (0.1\%) & 5.11785E-03 \\
triangle     & 1.88980E-03 (10.4\%) & 1.65986E-03 (3.0\%) & 1.71155E-03 \\
irregular & 4.49520E-03 (-)      & 4.20960E-03 (-)     & - \\
\end{tabular}%
}
\caption{Comparison of ANSYS, PINN, and analytical solutions with relative errors.}
\label{tab:summary}
\end{table}

\begin{figure}[h]
    \centering
    \includegraphics[width=\linewidth]{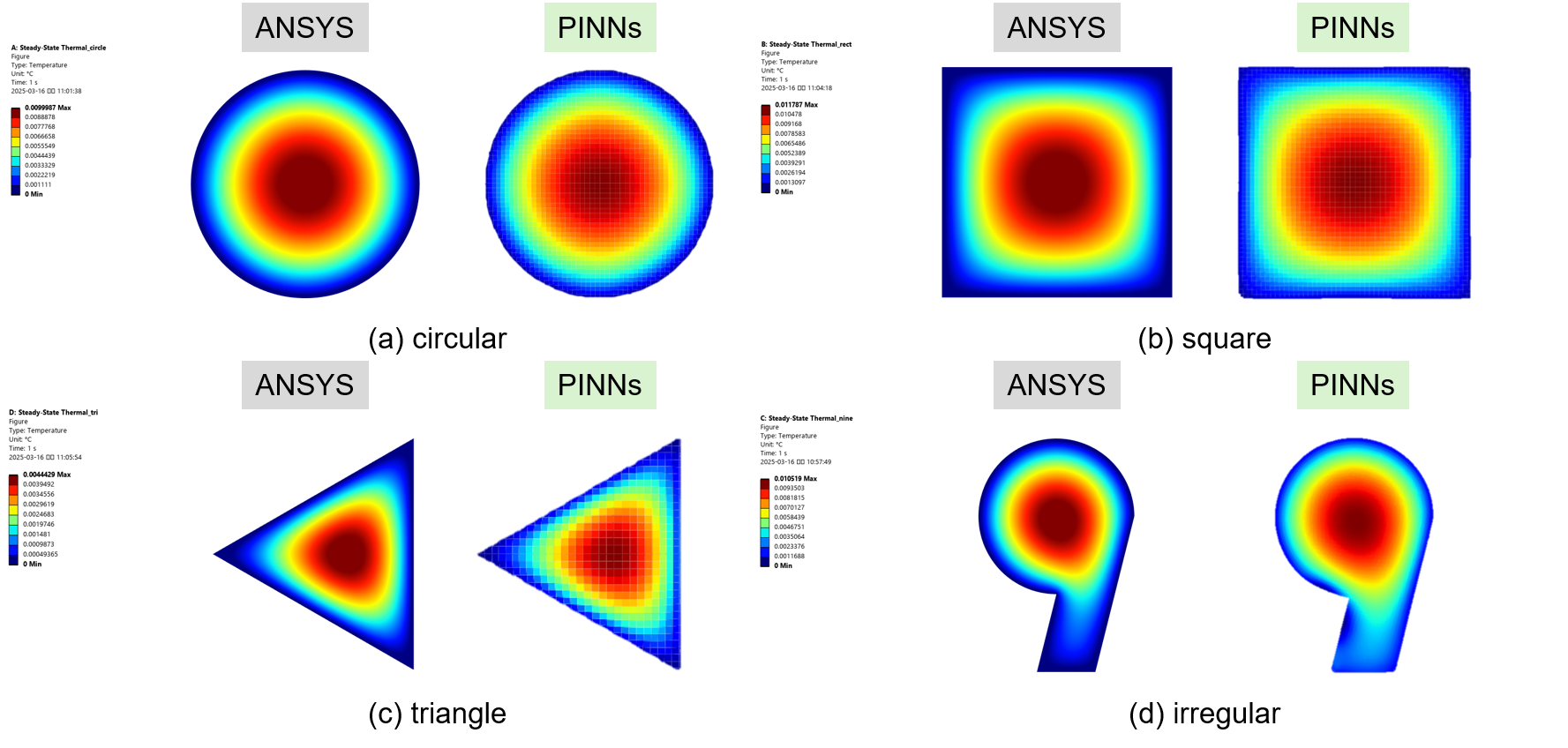}
    \caption{Comparison of Prandtl stress function field: ANSYS vs PINN.}
    \label{cant_comp}
\end{figure}

Figure \ref{cant_comp} compares the $\phi'$ contours obtained from ANSYS and PINN. In all cases, the PINN predictions closely resemble the ANSYS results, capturing the overall $\phi'$ distribution patterns and symmetry of the temperature fields. Despite slight differences in magnitude or boundary smoothness—especially for more complex shapes like the triangle and irregular geometry—the PINN effectively reproduces the $\phi'$ behavior. These results visually confirm the quantitative findings, demonstrating that PINN can achieve comparable accuracy to traditional solvers like ANSYS, even with complex or irregular domains.

\FloatBarrier 
\subsection{Single instance PINN
}

\begin{figure}[h]
    \centering
    \begin{subfigure}{0.4\textwidth}
        \centering
        \includegraphics[width=\linewidth]{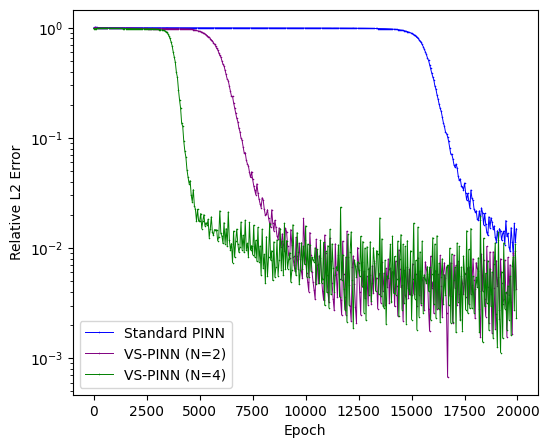}
        \caption{Learning curves for relative $L^2$-error}
    \end{subfigure}    
    \begin{subfigure}{0.40\textwidth}
        \centering
        \includegraphics[width=\linewidth]{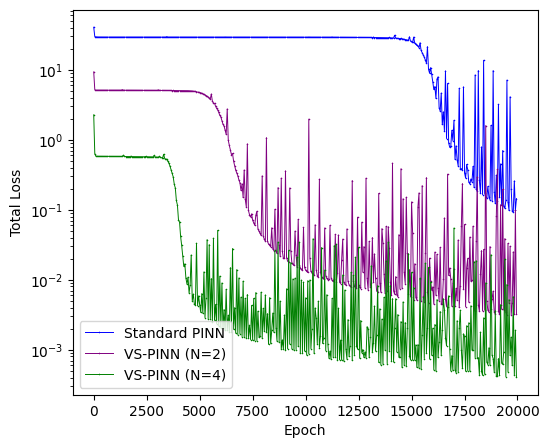}
        \caption{Learning curves for total loss}
    \end{subfigure}     
\caption{Training dynamics for the 1D torsion problem using VS-PINN with the scales $N=1$ (standard PINN), 2 and 4. (a) and (b) represent the relative $L^2$-errors and the total losses respectively, during training for each case.}
\label{vs_pinn_learn}
\end{figure}

\begin{figure}[h]
    \centering
    \begin{subfigure}{0.24\textwidth}
        \centering
        \includegraphics[width=\linewidth]{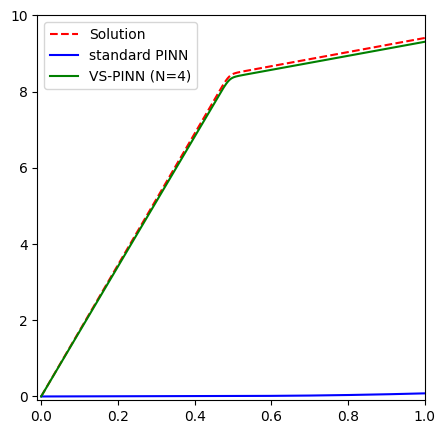}
        \caption{5500 epochs}
    \end{subfigure}    
    \begin{subfigure}{0.24\textwidth}
        \centering
        \includegraphics[width=\linewidth]{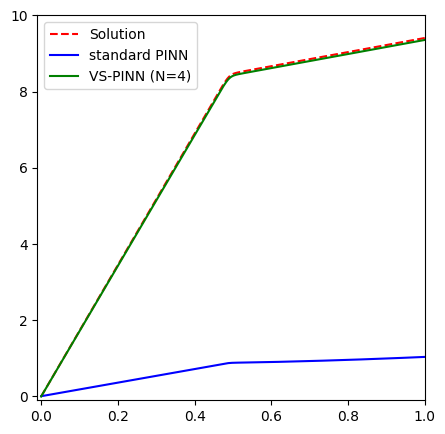} 
        \caption{15000 epochs}
    \end{subfigure}
    \begin{subfigure}{0.24\textwidth}
        \centering
        \includegraphics[width=\linewidth]{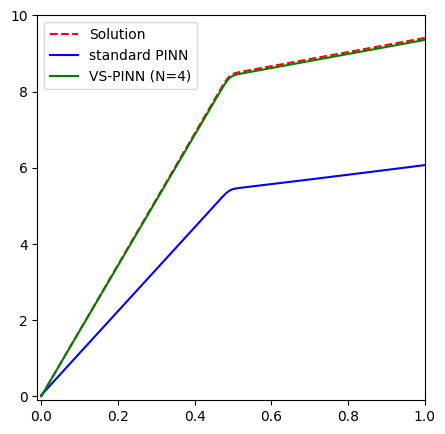}
        \caption{16000 epochs}
    \end{subfigure}    
    \begin{subfigure}{0.24\textwidth}
        \centering
        \includegraphics[width=\linewidth]{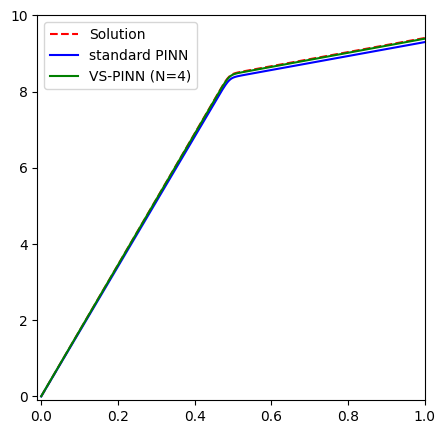}
        \caption{20000 epochs}
    \end{subfigure}    
\caption{Comparison of the solution predictions of VS-PINN and standard PINN after training up to a specified number of epochs.}
\label{VS-PINN_epoch}
\end{figure}
In this section, we will explore the case when the pipe's diameter has a sharp transition. When $J(x)$ changes rapidly, it may hinder the training of PINNs. To be more specific, we will solve the following problem:
\begin{equation}\label{1D_torsion}
\begin{aligned}
    \frac{{\rm{d}}}{{\rm{d}}x}\left(J\left(x\right)\frac{{\rm{d}}\phi}{{\rm{d}}x}\right)&= 0 \quad {\text{in }} [0,1],\\
    \phi(0)=0\quad&\text{and}\quad \phi'(1)=\frac{32T}{\pi GJ(1)}=\frac{32}{\pi J(1)}, 
\end{aligned}
\end{equation}
where $J(x)=(r^4(x)-(r(x)-0.2)^4)$ (the polar moment of inertia in a hollow circular cross-section) with $r(x)=1+{\rm{sigmoid}}(150x-75)$. To simplify the formula, we normalized the torsional force and shear modulus to 1. To handle the computational issues regarding the rapid transition of the coefficient in Eq. \eqref{1D_torsion}, we shall employ the VS-PINN introduced in Section \ref{sec:VS-PINN} to compute the solution. We will demonstrate that, in such cases, the VS-PINN can compute the solution more accurately and efficiently than the standard PINN. To do this, we shall compare the results obtained using VS-PINN with the scaling factor $N = 1$(standard PINN), $2$ and $4$. The corresponding scaled torsion bar problem can be written as
\begin{equation}\label{1D_torsion_equ}
\begin{aligned}
    N\frac{{\rm{d}}}{{\rm{d}}x}\left(J\left(\frac{x}{N}\right)N\frac{{\rm{d}}\phi}{{\rm{d}}x}\right)&= 0 \quad \text{in } [0,N],\\
    \phi(0)=0\quad&\text{and}\quad \phi'(N)=\frac{32}{\pi J(1)N},
\end{aligned}
\end{equation}
and we adopt the associated scaled loss as \[\mathcal{L}_{\rm scaled}=\frac{1}{N^4}\mathcal{L}_{\rm res}+20\mathcal{L}_{\rm data}.\]
We used the neural network with 2 hidden layers with 32 neurons per layer, and the hyperbolic tangent activation function. The model was trained with the Adam optimizer using 100 interior collocation data points, which were sampled from a uniform distribution over the computational domain. 

For a fair comparison, we used the same number of collocation points to train the standard PINN and the VS-PINN. As shown in Figure \ref{vs_pinn_learn}, increasing the scaling factor in the VS-PINN leads to a more rapid decrease in both relative $L^2$-error and total loss compared to the standard PINN.
After training the model up to 20,000 epochs, for the standard PINN, the highest accuracy achieved was the 9.7E$-3$ relative $L^2$-error. In contrast, when the model was trained with a variable scaling, the highest accuracy achieved was the 1.1E$-3$ relative $L^2$-error, showing that the VS-PINN outperforms the standard PINN when solving the problem with a sharp transition. Moreover, as can be seen in Figure \ref{VS-PINN_epoch}, the VS-PINN is able to predict the solution effectively with significantly fewer training epochs. Since the VS-PINN incurs no additional computational cost per epoch compared to the standard PINN, we can conclude that the VS-PINN achieves the solution both much faster and much more accurately than the standard PINN in this problem.

\subsection{Parametric PINN for surrogate model
}

\begin{figure}[h]
    \centering
    \begin{subfigure}{0.4\textwidth}
        \centering
        \includegraphics[width=\linewidth]{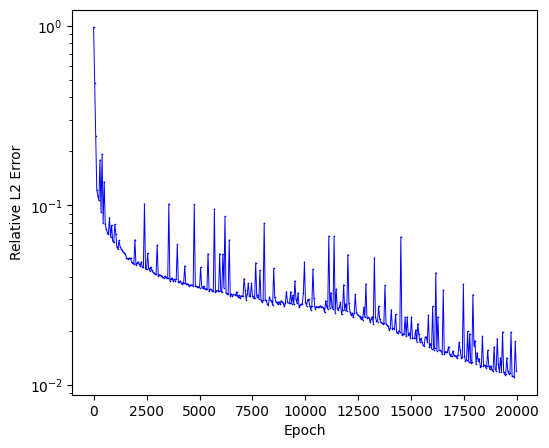}
        \caption{Error curve}
    \end{subfigure}    
    \begin{subfigure}{0.4\textwidth}
        \centering
        \includegraphics[width=\linewidth]{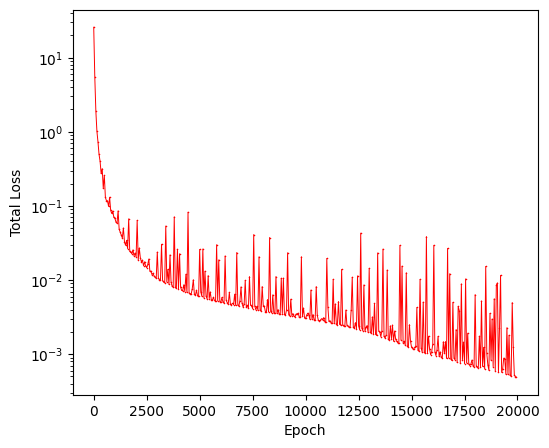}
        \caption{Loss curve}
    \end{subfigure} 
    \caption{(a) and (b) represent the relative $L^2$-errors and the total losses during training for the parametric PINN.}
    \label{para_train}
\end{figure}

As introduced in Section \ref{sec:Pa_PINN}, one of the key advantages of neural-network-based methods is that they can compute solutions to parametric PDE problems more efficiently. As described earlier, the parametric PINN is derived by integrating the conventional PINN loss function over the parameters that determine PDE data. In particular, we will train a model that can rapidly predict solutions as the mean, variance, and amplitude of a body force, given as a Gaussian distribution, change. To simplify the given situation, we set up the problem under the assumption that the outer and inner diameters of the pipe are constant and that the external forces follow a normal distribution. More precisely, for each $(T, m, \sigma)$, we shall consider the following parametric problem:
\begin{equation}\label{para_torsion_equ}
\begin{aligned}
    -\frac{{\rm{d}}^2\phi}{{\rm{d}}x^2}&= T\exp\bigg(-\frac{(x-m)^2}{2\sigma^2}\bigg) \quad \text{in } [0,1],\\
    \phi(0)&= \phi '(1)=0.
\end{aligned}
\end{equation}
Here, we incorporate the parameters $(T, m, \sigma) \in [1,10]\times[0.5,0.9]\times[0.2,1]$ defining the external force represented by a Gaussian distribution, as input variables to the neural network. With the notations introduced earlier in Section \ref{sec:Pa_PINN}, we employ a parametric loss function 
\[\mathcal{L}_{\rm para}=4\mathcal{L}_{{\rm{para}},\,r}+\mathcal{L}_{{\rm{para}},\,b}.\]
We trained a neural network with four hidden layers, each containing 64 neurons, and used the hyperbolic tangent activation function. For the training, we adopted the Adam optimizer and a number of sampling points $N_{pb}\times N_b=20000$, and $N_{pr}\times N_r=100000$, which were sampled from a uniform distribution. To evaluate accuracy, in this case, we measured error with respect to both the spatial variable $x\in[0,1]$ and the parametric variables $(T, m, \sigma) \in [1,10]\times[0.5,0.9]\times[0.2,1]$. After the training was completed, the highest accuracy was reported as 1.07E$-2$ relative $L^2$-error. The training dynamics can be found in Figure \ref{para_train}, where both the relative $L^2$-error and the total loss were measured during training. Note that in this case, the errors are measure with respect to both physical and parametric variables. 

\begin{figure}[h]
    \centering
    \begin{subfigure}{0.3\textwidth}
        \centering
        \includegraphics[width=\linewidth]{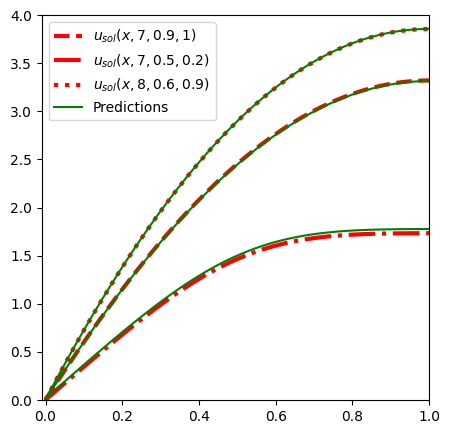}
        \caption{Set 1}
    \end{subfigure}   
    \begin{subfigure}{0.3\textwidth}
        \centering
        \includegraphics[width=\linewidth]{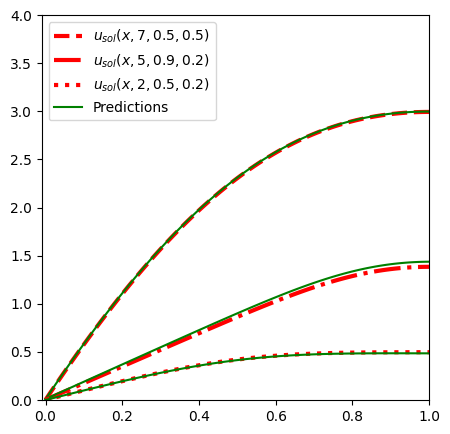} 
        \caption{Set 2}
    \end{subfigure}
    \begin{subfigure}{0.3\textwidth}
        \centering
        \includegraphics[width=\linewidth]{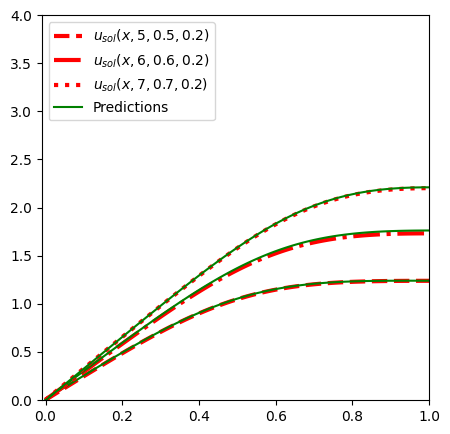} 
        \caption{Set 3}
    \end{subfigure}
    \caption{Comparison between the true solutions $u_{\rm sol}(x;T,m,\sigma)$ (red) and the predictions $\widehat{u}(x;T,m,\sigma)$ (green) obtained by the parametric PINN across various parameter sets $(T, m, \sigma)$.}
    \label{var_para}
\end{figure}

To further highlight the efficiency of our proposed method, we compare the predicted solutions of our model with the true solutions across various parameter configurations, as shown in Figure \ref{var_para}. 
More precisely, we randomly selected nine parameter sets from the parametric domain $[1,10]\times[0.5,0.9]\times[0.2,1]$, and computed the solutions for each using the parametric PINN model. 
For these nine parameter sets, the exact solutions and the corresponding predicted solutions were visualized in three groups, labeled as (a), (b), and (c) in the Figure \ref{var_para}. As we can see from the figure, our method is capable of accurately predicting the solutions even for a wide range of parameter values. As mentioned earlier, the key advantage of this methodology lies in its ability to provide real-time solution predictions as the PDE data (in this case, the external force) changes. This stands in clear contrast to the standard PINN, which require retraining the model whenever the PDE data is modified.

\section{Conclusion}
In this study, we proposed a series of PINN-based frameworks for solving Saint-Venant torsion problems, aiming to overcome the computational challenges inherent in conventional mesh-based numerical methods. We first demonstrated the successful application of the standard PINN to the 2D Poisson equation derived from the Prandtl stress function, achieving excellent agreement with analytical and FEM solutions across various geometries; notably, the relative errors were as low as 0.1\% for both circular and square shapes and 3.0\% for the triangular shape. To address cases involving sharp geometric transitions, a Variable-Scaling PINN (VS-PINN) was applied, reducing the relative $L^2$-error from 9.7E$-3$ (Vanilla PINN) to 1.1E$-3$, thereby significantly enhancing training efficiency and accuracy. Furthermore, we introduced a Parametric PINN capable of real-time surrogate modeling over a range of external force parameters without retraining, achieving a maximum relative $L^2$-error of 1.07E$-2$. These results collectively validate that PINN, VS-PINN, and parametric PINN offer user-specific demand solvers for torsional analysis in complex engineering structures, establishing a foundation for future extensions to broader elasticity problems.

\section*{Acknowledgments}
The authors would like to acknowledge the support
from the Hyundai Motor Group through the Future POC
program (Research Collaboration-Innovation
Technology for Vehicle) and the Nuclear Safety and Security Commission (NSSC) of the Republic of Korea (No. RS-2024-00403364). Seungchan Ko is supported by National Research Foundation of Korea Grant funded by the Korean Government (RS-2023-00212227).

\bibliographystyle{alpha}
\bibliography{sample}

\end{document}